\documentclass[
doubleblind,
]{ceurart}

\usepackage{listings}
\usepackage{url}            
\usepackage{nicefrac}       

\usepackage[textsize=tiny]{todonotes}

\usepackage{subfigure}
\usepackage{bm}

\usepackage{mathtools}
\usepackage{amsthm, thmtools}

\usepackage[capitalize,noabbrev]{cleveref}
\newcommand{\crefp}[1]{(\cref{#1})}
\Crefname{equation}{Eq.}{Eqs.}
\theoremstyle{plain}
\newtheorem{theorem}{Theorem}[section]

\theoremstyle{definition}

\newtheorem{assumption}[theorem]{Assumption}

\makeatletter
\def\th@remark{%
	\thm@headfont{\bfseries}%
	\normalfont 
	\thm@preskip\topsep \divide\thm@preskip\tw@
	\thm@postskip\thm@preskip
}
\makeatother
\theoremstyle{remark}

\usepackage{algorithm}
\usepackage{algpseudocode}
\usepackage{caption}

\usepackage{minitoc}
\renewcommand{\thepart}{}
\renewcommand{\partname}{}

\begin{document}
\doparttoc 
\faketableofcontents 


\copyrightclause{Copyright for this paper by its authors.
Use permitted under Creative Commons License Attribution 4.0
International (CC BY 4.0).}

\conference{
IAL@ECML-PKDD'24:
  8\textsuperscript{th} Intl. Worksh. \& Tutorial on Interactive Adaptive Learning,
  Sep. 9\textsuperscript{th}, 2024, Vilnius, Lithuania}
\title{Amortized Active Learning for Nonparametric Functions}

\author[1,2]{Cen-You Li}[
email=cen-you.li@campus.tu-berlin.de,
]
\address[1]{Technical University of Berlin, Germany}
\address[2]{Bosch Center for Artificial Intelligence, Germany}

\author[1]{Marc Toussaint}[%
email=toussaint@tu-berlin.de,
]

\author[2]{Barbara Rakitsch}[%
email=barbara.rakitsch@de.bosch.com,
]
\fnmark[1]

\author[2]{Christoph Zimmer}[%
email=christoph.zimmer@de.bosch.com,
]
\fnmark[1]

\cortext[1]{Corresponding author.}
\fntext[1]{equal contribution}

\begin{abstract}
Active learning (AL) is a sequential learning scheme aiming to select the most informative data.
AL reduces data consumption and avoids the cost of labeling large amounts of data.
However, AL trains the model and solves an acquisition optimization for each selection.
It becomes expensive when the model training or acquisition optimization is challenging.
In this paper, we focus on active nonparametric function learning, where the gold standard Gaussian process (GP) approaches suffer from cubic time complexity.
We propose an amortized AL method, where new data are suggested by a neural network which is trained up-front without any real data~\crefp{figure1}.
Our method avoids repeated model training and requires no acquisition optimization during the AL deployment.
We (i) utilize GPs as function priors to construct an AL simulator, (ii) train an AL policy that can zero-shot generalize from simulation to real learning problems of nonparametric functions and (iii) achieve real-time data selection and comparable learning performances to time-consuming baseline methods.
\end{abstract}

\maketitle

\section{Introduction}\label{section1-introduction}

Active learning (AL) is a sequential learning scheme aiming to reduce the effort and cost of labeling data~\citep{settles2010_al,KumarGupta2020,tharwat2023_al}.
The goal is to maximize the information given by each data point, so the quantity can be reduced.
An Active Learning (AL) method starts with a small amount of labeled data.
The model is first trained on the labeled data, and then the trained model is used to evaluate acquisition scores for the unlabeled data.
The acquisition function measures the expected knowledge gained from labeling a data point.
Labels are then requested for the data points with the peaked acquisition scores, and the labeled dataset is updated for the next AL iteration.
AL can be run for several iterations until the budget is exhausted or until a training goal is achieved.
To perform AL, however, one would face multiple challenges:
(i) training models for every query can be nontrivial, especially when the learning time is constrained~\citep{Gal2017_icml_bal,kirsch2019batchbald,Lederer2021_icml_gprealtimecontrol};
(ii) acquisition criteria need to be selected a priori but none of them clearly outperforms the others in all cases, which makes the selection difficult~\citep{Baram2004_jmlr_onlineAL,Konyushkova2017_learnAL};
(iii) optimizing an acquisition function can be difficult (e.g. sophisticated discrete search space~\cite{Swersky2020amortized_bo}).

\begin{figure}[t]
	\vskip 0.2in
	\begin{center}
        \centerline{\includegraphics[width=\textwidth]{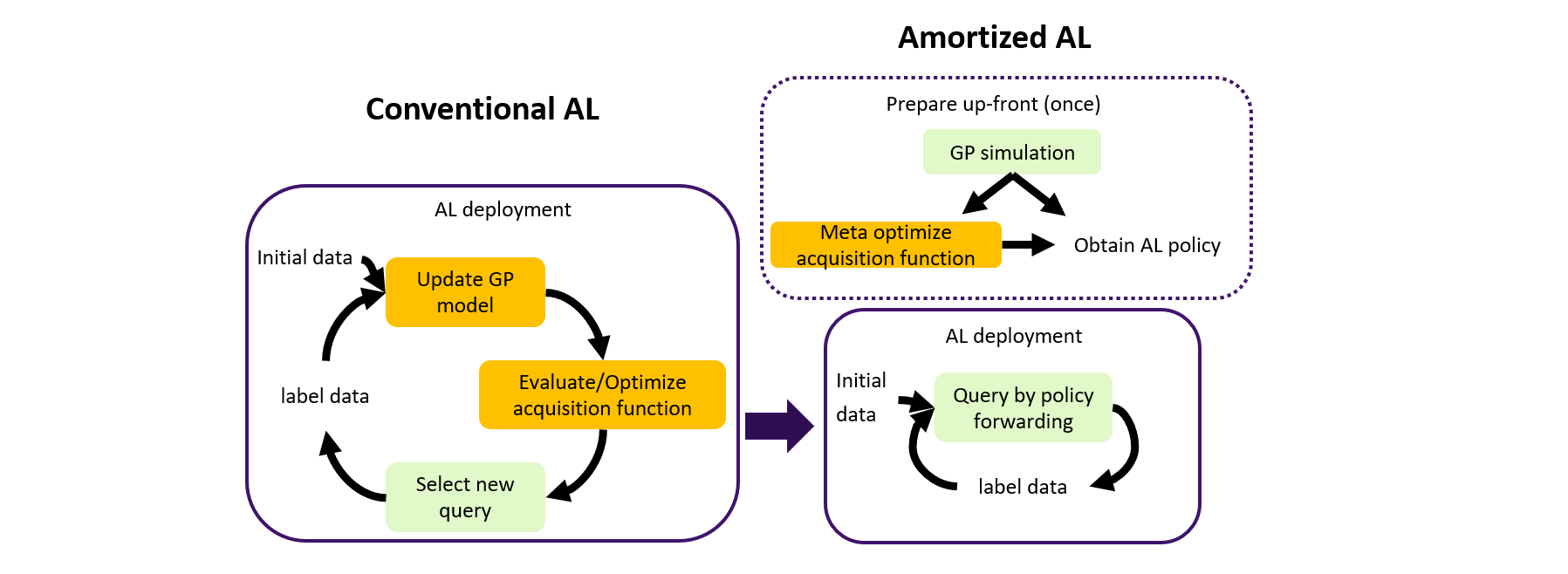}}
        \captionof{figure}{The conventional AL of a GP regression relies on computationally expensive (orange) GP fitting and acquisition optimization.
        Our amortized AL approach meta trains a NN active learner up-front, purely from synthetic data, allowing a fast, easy and real-time applicable (green) AL deployment.
        }
		\label{figure1}
	\end{center}
\vskip -0.2in
\end{figure}

In this paper, we propose an AL method that suggests new data points for labeling based on a neural network (NN) evaluation instead of the costly model training and acquisition function optimization~\crefp{figure1}.
To this end, we decouple model training and acquisition function optimization from the AL loop. 
This is beneficial when we face the aforementioned challenges (i) and (iii), i.e. scenarios where either the querying time (training time pluses optimization time) is precious~\citep{Gal2017_icml_bal,kirsch2019batchbald,Lederer2021_icml_gprealtimecontrol} or it is difficult to optimize an acquisition function~\citep{Swersky2020amortized_bo}.
In these settings, making a high-quality data selection is too expensive, such that one would rather accept a faster and easier active learner even with a potential tradeoff of slightly worse acquisition quality.
Notably, as AL tackles data scarcity problem, such a NN policy function should be obtained with no additional real data.

We further focus our problem on actively learning regression tasks.
The idea is to
(i) generate a rich distribution of functions,
(ii) simulate AL experiments on those functions,
(iii) train the NN policy in simulation, and then
(iv) zero-shot generalize to real AL problems.
For low data learning problems (up to thousands of data points), Gaussian processes (GPs,~\cite{GPbook}) are a powerful model family that naturally fits our approach.
A GP is a distribution of nonparametric functions that, if used as the model in an AL loop (\cref{figure1}, left), provides well-calibrated predictive distributions suitable for acquisition functions~\citep{guestrin_mi_al_05,krause_nonmyopic_al_07,krause08a,houlsby_2011_BALD}.
This paper utilizes GPs to sample functions and simulate AL of regression problems (\cref{figure1}, right).
In other words, we perform amortized inference~\citep{Gershman2014amortized_inference} of an active learner from GP simulations.

Please notice the difference between the model and the NN policy.
In this paper, model always refers to the model one wish to actively learn on a specific task, while the NN policy proposes AL queries and the queries are then used to fit the model.

\paragraph{Contributions:}
We summarize our contributions:
\begin{itemize}
	\item we formulate a training pipeline of active nonparametric function learning policy which requires no real data;
	\item we propose differentiable AL objectives in closed form for the training;
	\item we demonstrate empirical analysis on common benchmark problems.
\end{itemize}

\paragraph{Related works:}\label{section1.2-related_works}

\begin{wrapfigure}{R}{0.5\textwidth}
	\vspace{-5pt}
	\begin{minipage}{\linewidth}
		\captionof{algorithm}{Classical AL}
		\label{alg-classical_al}
		\begin{algorithmic}[1]
			\Require $\mathcal{D}_0 \subseteq \mathcal{X}\times\mathcal{Y}$, acquisition function $a$
			\For{$t=1, ..., T$}
			\State Model $\mathcal{M}_{t-1}$ with $\mathcal{D}_{t-1}$
			\State $\bm{x}_t = argmax_{\bm{x}\in\mathcal{X}} a(x | \mathcal{M}_{t-1}, \mathcal{D}_{t-1})$
			\State Evaluate $y_t$ at $\bm{x}_t$
			\State $\mathcal{D}_{t} \gets \mathcal{D}_{t-1} \cup \{ \bm{x}_t, y_t \}$
			\EndFor
		\end{algorithmic}
	\end{minipage}
	\vspace{-5pt}
\end{wrapfigure}

AL~\citep{settles2010_al,KumarGupta2020,tharwat2023_al} is prominent in various applications such as image classification~\citep{Gal2017_icml_bal,kirsch2019batchbald} or physical system modeling~\citep{ZimmerNEURIPS2018_b197ffde}.
In regression tasks, GPs demonstrate great advantage in AL acquisitions~\citep{GarnettOH2013,Schreiter2015,ZimmerNEURIPS2018_b197ffde,Yue2021_al_gp_shape_control,cyli2022,bitzer23a_nonstationary_al}.
An acquisition function plays a major role in AL methods~\crefp{alg-classical_al}.
Entropy, which selects the most uncertain points in the space, is a popular acquisition function due to its effectiveness and computational simplicity~\citep{Seo2000GaussianPR}.
Mutual information is another well-known option.
A mutual information criterion can focus on the information gain in the space~\citep{guestrin_mi_al_05,krause08a} or take model improvement~\citep{houlsby_2011_BALD} into account, which is often considered superior to entropy.
However, depending on the settings, mutual information is often intractable and creates computational overhead.
A closely related field is Bayesian optimization (BO,~\cite{brochu2010tutorial}) which aims to find the global optima of functions with limited evaluations.
The same algorithm~\crefp{alg-classical_al} can be applied to BO problem by exchanging the acquisition function.
BO as well suffers from repeated model training and acquisition optimization.

Recently, meta learning and amortized inference have been explored to tackle challenges of sequential learning methods.
Konyushkova et al. proposed to meta learn an acquisition function for AL, avoiding a priori selection~\citep{Konyushkova2017_learnAL}.
Given an acquisition function, Swersky et al. proposed to do an amortized inference on acquisition optimization~\citep{Swersky2020amortized_bo}.
On GP learning problems, Rothfuss et al. proposed to meta learn GP hyperparameters~\citep{rothfuss21pacoh} while Bitzer et al. performed amortized inference to select GP kernels and hyperparameters~\cite{Bitzer2023amortized_gp}, both of which simplify the model fitting which is a bottleneck in real time applications~\citep{Lederer2021_icml_gprealtimecontrol}.

To the best of our knowledge, very rare works automate the entire data selection process, i.e. decouple model updates, automate acquisition evaluations and optimizations.
In~\cite{Andrychowicz2016_learn_to_learn,Chen2017_learn_to_learn}, the authors proposed RNN optimizers which query points by simple forwarding.
Foster et al. proposed the deep adaptive design (DAD), an amortized Bayesian experimental design, which as well resorts sequential data selection to simple NN forwarding~\citep{foster_dad_2021}.
While DAD provides an AL deployment procedure as we aim for, they collect data to learn parametric models.
The data selection criterion does not necessarily fit into nonparametric functions.
Ivanova et al. further extended DAD to learn intractable models, which is however a different direction from our goal~\citep{ivanova_idad_2021}.

None of the literature we found considers amortized inference of active nonparametric function learning.
Interestingly, Krause et al. discussed theoretical perspectives of an a priori acquisition policy for active GP learning~\citep{krause_nonmyopic_al_07}.
This provides key insight into our AL simulation.
We take inspiration from~\cite{foster_dad_2021,Chen2017_learn_to_learn,krause_nonmyopic_al_07} to develop our amortized AL method.

\section{Problem Statement}
\begin{wrapfigure}{R}{0.5\textwidth}
	\vspace{-5pt}
	\begin{minipage}{\linewidth}
		\captionof{algorithm}{AL with NN Policy}
		\label{alg-policy_al}
		\begin{algorithmic}[1]
			\Require $\mathcal{D}_0\subseteq \mathcal{X}\times\mathcal{Y}$, AL policy $\phi$
			\For{$t=1, ..., T$}
			\State $\bm{x}_t = \phi(\mathcal{D}_{t-1})$
			\State Evaluate $y_t$ at $\bm{x}_t$
			\State $\mathcal{D}_{t} \gets \mathcal{D}_{t-1} \cup \{ \bm{x}_t, y_t \}$
			\EndFor
			\State Model $f$ with $\mathcal{D}_{T}$
		\end{algorithmic}
	\end{minipage}
	\vspace{-5pt}
\end{wrapfigure}

We are interested in a regression task of an unknown function $f: \mathcal{X} \rightarrow \mathbb{R}$, where $\mathcal{X} \subseteq \mathbb{R}^D$ is the input space.
We assume $\mathcal{X}$ is a bounded space, which usually holds true in reality, as one normally focuses only on a domain of interest.
The observations we access are always noisy.
That is, a labeled data point comprises an input $\bm{x} \in \mathcal{X}$ and its corresponding output observation $y(\bm{x})=f(\bm{x}) + \epsilon$, where $f(\bm{x})$ is a functional value and $\epsilon$ is an unknown noise value.
For brevity, we write $y \coloneqq y(\bm{x})$ and $y_{subscript} \coloneqq y(\bm{x}_{subscript})$.
For clarity later, we let $\mathcal{Y} \subseteq \mathbb{R}$ denote the output space, i.e. $y \in \mathcal{Y}$, $\mathcal{D} \subseteq \mathcal{X}\times\mathcal{Y}$ denote a dataset, and $space(\mathcal{X} \times \mathcal{Y}) \coloneqq \{ \mathcal{D} | \mathcal{D} \subseteq \mathcal{X} \times \mathcal{Y} \}$ denote the space of datasets.

We follow an AL setting:
a small labeled dataset $\mathcal{D}_0 \coloneqq \{ \bm{x}_{init, i}, y_{init, i} \}_{i=1}^{N_{init}}$ is given, and we have budget to label $T$ more data points, denoted by $(\bm{x}_1, y_1),..., (\bm{x}_T, y_T)$.
The high level goal is to conduct AL to select informative $\bm{x}_1, ..., \bm{x}_T$ such that $\mathcal{D}_T=\mathcal{D}_0 \cup \{ \bm{x}_1, y_1, ..., \bm{x}_T, y_T \}$ helps us construct a good model of $f$.
In a conventional AL method (\cref{figure1}, left and~\cref{alg-classical_al}), the data are selected iteratively by optimizing the acquisition criteria.
In this paper, we aim to have a policy function $\phi: space(\mathcal{X} \times \mathcal{Y}) \rightarrow \mathcal{X}$ up front, which sees current observations and directly provide the next query proposal (\cref{figure1}, right and~\cref{alg-policy_al}).
We assume no additional real data are available for the policy training.
Nevertheless, we make assumptions that $f$ has a GP prior and that our observation data are normalized to zero mean and unit variance.
In the following, we will sometimes write $\bm{X}_{init} \coloneqq (\bm{x}_{init,1},...,\bm{x}_{init,N_{init}})$,
$Y_{init} \coloneqq (y_{init,1},...,y_{init,N_{init}})$, $\bm{X}_t \coloneqq (\bm{x}_{init,1},...,\bm{x}_{init,N_{init}}, \bm{x}_1, ..., \bm{x}_t)$,
$Y_t \coloneqq (y_{init,1},...,y_{init,N_{init}}, y_1, ..., y_t)$,
for $t=1, ..., T$.

\paragraph{Assumptions:}

We assume $f$ has a GP prior.
A GP is a distribution over functions, characterized by the mean ($\mathbb{E}[f(\bm{x})]$) and kernel (covariance $f(\bm{x})$ and $f(\bm{x}')$, for two input points $\bm{x}$, $\bm{x}'$).
Without loss of generality, one usually assumes the mean is a zero function, which holds true when the observation values are normalized.
The kernel function is typically parameterized, and it encodes the amplitude and smoothness of the function $f$.
We refer the readers to~\cite{GPbook} for details.
The assumption is formally described below.
\begin{assumption}\label{assump-gp_prior}
	The unknown function $f$ has a GP prior $\mathcal{GP}(\bm{0}, k_\theta)$.
	Any observation at $\bm{x}$ is $y = f(\bm{x}) + \epsilon$, $\epsilon \sim \mathcal{N}(0, \sigma^2)$ is an i.i.d. Gaussian noise~\citep{GPbook}.
	Here, $k_\theta: \mathcal{X} \times \mathcal{X} \rightarrow \mathbb{R}$ is a kernel function parameterized by $\theta$.
	We further assume $k_\theta(\bm{x}, \bm{x}') \leq 1$.
\end{assumption}
Bounding the kernel scale by one is not restrictive, as we assume the observations are normalized to unit variance.
Due to a GP prior, any finite number of functional values are jointly Gaussian.
GP distributions are provided in closed form in~\cref{sectionS2-gp_details}.

We want to emphasize that the GP assumption is mainly for policy training.
On a test function, failing this assumption (we however would not know a priori) may result in bad data selection, but our AL method can still be deployed as the data selection is decoupled from GP modeling.

\section{AL with a priori trained policy $\phi$}\label{section3-method}

\begin{wrapfigure}{R}{0.5\textwidth}
\vspace{-5pt}
\begin{minipage}{\linewidth}
\captionof{algorithm}{Nonmyopic AL training}
	\label{alg-nonmyopic_al_training}
	\begin{algorithmic}[1]
		\Require prior $\mathcal{GP}(0, k_\theta), p(\epsilon)=\mathcal{N}(0, \sigma^2)$, $T$
		\State sample a batch of $\theta, \sigma^2$
		\State sample a batch of $f \sim \mathcal{GP}(0, k_\theta)$
		\State sample $\mathcal{D}_0 \subseteq \mathcal{X}\times\mathcal{Y}$, given $f$ and $p(\epsilon)$
		\For{$t=1, ..., T$}
		\State $\bm{x}_{t} = \phi(\mathcal{D}_{t-1})$
		\State sample $\epsilon_{t} \sim p(\epsilon), y_{t}=f(\bm{x}_{t}) + \epsilon_t$
		\State $\mathcal{D}_{t} \gets \mathcal{D}_{t-1} \cup \{ \bm{x}_{t}, y_{t} \}$
		\EndFor
		\If{entropy loss}
		\State compute loss per~\cref{eq-gp_entropy_loss2_with_initial_points}
		\ElsIf{regularized entropy loss}
		\State sample $\bm{X}_{grid} \subseteq \mathcal{X}$
		\State sample $Y_{grid}=f(\bm{X}_{grid}) + noise$
		\State compute loss per~\cref{eq-gp_mi_loss2_grid_approx_with_initial_points}
		\EndIf
        \State update $\phi$
	\end{algorithmic}
\end{minipage}
\vspace{-5pt}
\end{wrapfigure}

Our goal here is to train a policy $\phi$ to run~\cref{alg-policy_al}.
Here we take key inspiration from~\cite{foster_dad_2021,ivanova_idad_2021}.
The idea is to exploit the GP prior~\crefp{assump-gp_prior} before AL experiments.
We use the GP prior $p(f)$ and the Gaussian likelihood
$p(y | \bm{x}, f)=\mathcal{N}\left(
y | f(\bm{x}), \sigma^2
\right)$
to construct a simulator.
This allows us to sample functions, simulate policy-based AL~\crefp{alg-policy_al} and then meta optimize an objective function which encodes the acquisition criterion~\crefp{alg-nonmyopic_al_training}.
The key is to ensure that the policy experiences AL on diverse functions, then during a real AL experiment, the policy makes a zero-shot amortized inference from the simulation.
Note that the training is performed by simulating active GP learning, while, in a real AL experiment, the policy only collects data, and we are not forced to make GP modeling with the collected data.

\paragraph{Training objectives:}
We first discuss the training objectives, as they provide insight into what exact data we generate.
Similar to~\cite{Chen2017_learn_to_learn,foster_dad_2021}, the idea is to turn the acquisition criteria we would have optimized in a conventional AL setting into loss objectives where the learner gradient is available~\crefp{figure1}.

Imagine we are doing AL with~\cref{alg-policy_al} on synthetic functions.
The first remark is that in a simulation, functions are always sampled from a known GP prior, i.e. parameters $\theta, \sigma^2$ are known before we start the simulated AL procedure.
Thus, given a sequence of queries provided by a learner, the joint GP distribution is available in closed form.
Therefore, an intuitive approach is to apply common entropy or (approximated) mutual information criteria on the policy selected points.
We take the definition from~\cite{krause_nonmyopic_al_07}, where the authors discuss policy-based AL which naturally applies to NN policies as well:
\begin{align}
\label{eq-gp_entropy_loss2}
\mathcal{H}(\phi) &\coloneqq
\mathbb{E}_{
	p(f(\cdot), \epsilon_{t=1, ...,T})
}
\left[
-\log p(y_{\phi,1}, ..., y_{\phi,T})
\right],\\
\label{eq-gp_mi_loss2}
\mathcal{I}(\phi) &\coloneqq
\mathbb{E}_{
	p(f(\cdot), \epsilon_{t=1, ...,T})
} \left[
-\log p(y_{\phi,1}, ..., y_{\phi,T}) + \log p( y_{\phi,1}, ..., y_{\phi,T} | y(\mathcal{X} \setminus \bm{X}_\phi) )
\right],
\end{align}
where $f(\cdot)$ and $\epsilon_{t=1, ...,T}$ are GP and noise realizations, $y_{\phi,1}, ..., y_{\phi,T}$ correspond to policy selected queries $\bm{x}_{\phi,1}, ..., \bm{x}_{\phi,T}$, and $y(\mathcal{X} \setminus \bm{X}_\phi)$ means the realization over space $\mathcal{X} \setminus \{ \bm{x}_{\phi,1}, ..., \bm{x}_{\phi,T} \}$.
In~\cite{krause_nonmyopic_al_07}, the input space $\mathcal{X}$ is a discrete space of finite number of elements, which makes $y(\mathcal{X} \setminus \bm{X}_\phi)$ a computable set of values.
We will describe $y(\mathcal{X} \setminus \bm{X}_\phi)$ in more details later. 
Note here that stochasticity arises from the function sampling while the AL policy is dealing with each function deterministically.

Maximizing the entropy objective~\crefp{eq-gp_entropy_loss2} would favor a set of uncorrelated points and naturally encourage points at the border which are the most scattered~\citep{guestrin_mi_al_05}.
In our initial experiments, we noticed that this entropy objective needed more careful tuning, as it often overemphasized the boundary and ignored to explore in the space.
The mutual information criterion is known to tackle this problem, at least in conventional AL settings~\citep{guestrin_mi_al_05}, but, on the other hand, the aforementioned objective $\mathcal{I}(\phi)$ in its original form makes conditioning on $y(\mathcal{X} \setminus \bm{X}_\phi)$.
This is not well-defined when $\mathcal{X}$ is a continuous space.
Even if $\mathcal{X}$ is discrete, conditioning on large pool (fine discretization) is computationally heavy, i.e. GP cubic complexity $\mathcal{O}\left( |\mathcal{X}|^3 \right)$~\crefp{sectionS2-gp_details}.
Discrete pool also enforces a classifier-like policy $\phi$ (select points from a pool) which prohibits us from utilizing the existing NN structure developed by~\cite{foster_dad_2021,ivanova_idad_2021}.

We thus wish to modify $\mathcal{I}(\phi)$.
Note that $\mathcal{I}(\phi)$ is a regularized entropy objective, and $\mathcal{H}(\phi)$, although not always well performing, can already be used for training.
Therefore, we propose a simple yet effective approach: compute the regularization term only on a sparse set of $N_{grid}$ samples $\left( \bm{X}_{grid}, Y_{grid} \right) \in space(\mathcal{X} \times \mathcal{Y})$, i.e.
\begin{align}\label{eq-gp_mi_loss2_grid_approx}
\mathcal{I}(\phi) \approx \mathbb{E}_{
	p(f(\cdot), \epsilon_{t=1, ...,T})
} \left[
-\log p(y_{\phi,1}, ..., y_{\phi,T}) + \log p( y_{\phi,1}, ..., y_{\phi,T} | Y_{grid} )
\right].
\end{align}
$N_{grid}$ should be much larger than $T$.
Maximizing this objective encourages $\{ \bm{x}_{\phi,1},...,\bm{x}_{\phi,T} \}$ to track subsets of $\bm{X}_{grid}$.
To keep the policy from selecting only those sparse grid samples, which are not necessarily optimal points, we re-sample $\bm{X}_{grid}$ in each training step.
The intuition of this objective is two-fold:
(i) it can be viewed as an entropy objective regularized by an additional search space indicator, or
(ii) it can be viewed as an imitation objective because a subset of grid points, if happens to have large joint entropy, maximizes the objective.

The above losses consider a fixed set of GP hyperparameters, which encodes only certain function features.
To generalize to diverse functions, we take the GP hyperparameters into account, and note that a real AL is initiated with initial data points.
Our policy objectives become
\begin{align}
\begin{split}
\label{eq-gp_entropy_loss2_with_initial_points}
\mathcal{H}(\phi) & =
\mathbb{E}_{
	p(\theta, \sigma^2)
}
\mathbb{E}_{
	p(f(\cdot), \epsilon_{t=1, ...,T})
}
\left[
-\log p(y_{\phi,1}, ..., y_{\phi,T}, Y_{init})
\right]\\
&\propto \mathbb{E}_{
	p(\theta, \sigma^2)
}
\mathbb{E}_{
	p(f(\cdot), \epsilon_{t=1, ...,T})
}
\left[
-\log p(y_{\phi,1}, ..., y_{\phi,T} | Y_{init})
\right]
\end{split}
\end{align}
\begin{align}
\begin{split}
\label{eq-gp_mi_loss2_grid_approx_with_initial_points}
\mathcal{I}(\phi)
&\approx \mathbb{E}_{
	p(\theta, \sigma^2)
}
\mathbb{E}_{
	p(f(\cdot), \epsilon_{t=1, ...,T})
} \left[
-\log p(y_{\phi,1}, ..., y_{\phi,T}, Y_{init})
+\log p( y_{\phi,1}, ..., y_{\phi,T}, Y_{init} | Y_{grid} )
\right]\\
&\propto \mathbb{E}_{
	p(\theta, \sigma^2)
}
\mathbb{E}_{
	p(f(\cdot), \epsilon_{t=1, ...,T})
} \left[
-\log p(y_{\phi,1}, ..., y_{\phi,T} | Y_{init})
+\log p( y_{\phi,1}, ..., y_{\phi,T} | Y_{init}, Y_{grid} )
\right].
\end{split}
\end{align}
The proportion symbol here indicates equivalency, and this holds by applying Bayes rule and removing the part that is not relevant to the policy gradient.
In this paper, we sample $\bm{X}_{grid}, \theta, \sigma^2$ uniformly.
Please see the appendix for numerical details.

To summarize, we are given priors $p(f) \sim \mathcal{GP}(0, k_{\theta}), p(\epsilon) \sim \mathcal{N}(0, \sigma^2)$ with uniformly random hyperparameters $\theta$ and $\sigma^2$, we may then sample GP function and noise realizations and a policy returns sequences of data by actively learning those functions.
Then the data are plugged into meta AL objectives~\crefp{eq-gp_entropy_loss2_with_initial_points,eq-gp_mi_loss2_grid_approx_with_initial_points} where the gradient propagates from the queries backward into the policy.
We see here that data are generated where~\cref{eq-gp_entropy_loss2_with_initial_points,eq-gp_mi_loss2_grid_approx_with_initial_points} require, and this allows one to easily sample thousands or millions of functions in the training.
In the next section, we zoom into the sampling of the policy-queried data.

\paragraph{Simulated AL:}

The objective functions provide insight into the simulation procedure:
sample a GP function realization,
sample initial data,
perform AL cycles by forwarding with the policy, and maximize either the policy entropy~\crefp{eq-gp_entropy_loss2_with_initial_points} or the regularized policy entropy (i.e. the modified mutual information~\cref{eq-gp_mi_loss2_grid_approx_with_initial_points}).
The training procedure is summarized in~\cref{alg-nonmyopic_al_training}.
One can see that lines 4-8 are simulating AL cycles as how the policy will be deployed~\crefp{alg-policy_al}.

The only remaining challenge here is to ensure that $y_{\phi,1}, ..., y_{\phi,T}$ are from the same GP function.
This is not trivial because the observations are sampled iteratively, i.e. $\forall t=1,...,T$, $\bm{x}_t=\phi(\mathcal{D}_{t-1})$, which means $y_{\phi,1}, ..., y_{\phi,t-1}$ need to be sampled before $\bm{x}_t,..., \bm{x}_T$ are known.
One way is to make a standard GP posterior sampling $y_t \sim p(y(\bm{x}_t) | \mathcal{D}_{t-1}, k_\theta, \sigma^2)$ instead of line 2 and 6 of~\cref{alg-nonmyopic_al_training}.
However, this results in $\mathcal{O}\left(N_{init}^3 + (N_{init} + 1)^3 + ... + (N_{init} + T-1)^3\right)$ complexity in time, i.e. the notorious GP cubic complexity~\crefp{sectionS2-gp_details}.
Sampling $Y_{grid}$ (line 11 of~\cref{alg-nonmyopic_al_training}) would also take tremendous time.

We address this issue by applying a decoupled function sampling technique~\citep{Rahimi2007_rff,Wilson2020_icml_rff_gp_post}.
The idea is to sample Fourier features to approximate a GP function.
As a result, an approximated function is a linear combination of cosine functions (line 2 of~\cref{alg-nonmyopic_al_training}), and we can later compute the function value at any point $\bm{x}\in\mathcal{X}$ in linear time (line 6 \& 13 of~\cref{alg-nonmyopic_al_training}).
One limitation arises, however, is that the kernel $k_{\theta}$ needs to have a Fourier transform (e.g. stationary kernels, see Bochner's theorem in~\cite{GPbook}).

Notice that this training procedure simulates a nonmyopic AL.
That is, the $T$ queries are optimized if considered jointly but not necessarily stepwise optimal.
We additionally provide a myopic AL training algorithm detailed in appendix~\crefp{alg-myopic_al_training}, which optimizes stepwise data selection.
This idea is simple: the initial dataset has size randomly sampled from $\{ N_{init}, ..., N_{init} + T-1 \}$, the policy query one point, and then we compute the same loss objectives with the altered sequential structure.
A myopic policy is not expected to have better AL performance but can avoid making recursive NN inference during the training.
This might be beneficial if we want to scale the training up to larger $N_{init}$ or larger $T$.

\paragraph{NN structure:}
We take the NN described in~\cite{ivanova_idad_2021}.
Each data pair $(\bm{x}, y)$ is first mapped by a MLP (multilayer perceptrons) to a $32$ dimension embedding, then two layers of transformer encoders (\citep{Vaswani2017_attention}, without positional encoding) are applied to the sequence of data pair embeddings, and finally the attended sequence is summed (ensure permutation invariance of observed dataset) before mapped by another MLP to a new data query.
The details are described in~\cite{foster_dad_2021,ivanova_idad_2021}.
We only add another Tanh layer with rescaling constants to refine the decoder output (refine the output to $\mathcal{X}$ which is bounded in our case).
The query is in continuous $\mathcal{X}$ space and this is how we train the policy.
If an AL problem is considered over discrete $\mathcal{X}$, one simple approach is to select the point closest to the NN query (line 2 of~\cref{alg-policy_al}).

\paragraph{Complexity:}
The training complexities are in~\cref{sectionS4-training_complexity}.
The AL deployment complexities are as follows
\begin{itemize}
	\item amortized AL~\crefp{alg-policy_al}: NN forwarding takes $\mathcal{O}\left( (N_{init}+t-1)^2 \right)$ at each $t$;
	\item conventional GP AL~\crefp{alg-classical_al}: at each $t$, GP modeling takes $\mathcal{O}\left( (N_{init}+t-1)^3 \right)$ in time while complexity of acquisition optimization depends on the exact AL problems.
\end{itemize}

\section{Experiments}\label{section4-experiments}

\begin{figure}[t]
	\vskip 0.2in
	\begin{center}
		\centerline{\includegraphics[width=\textwidth]{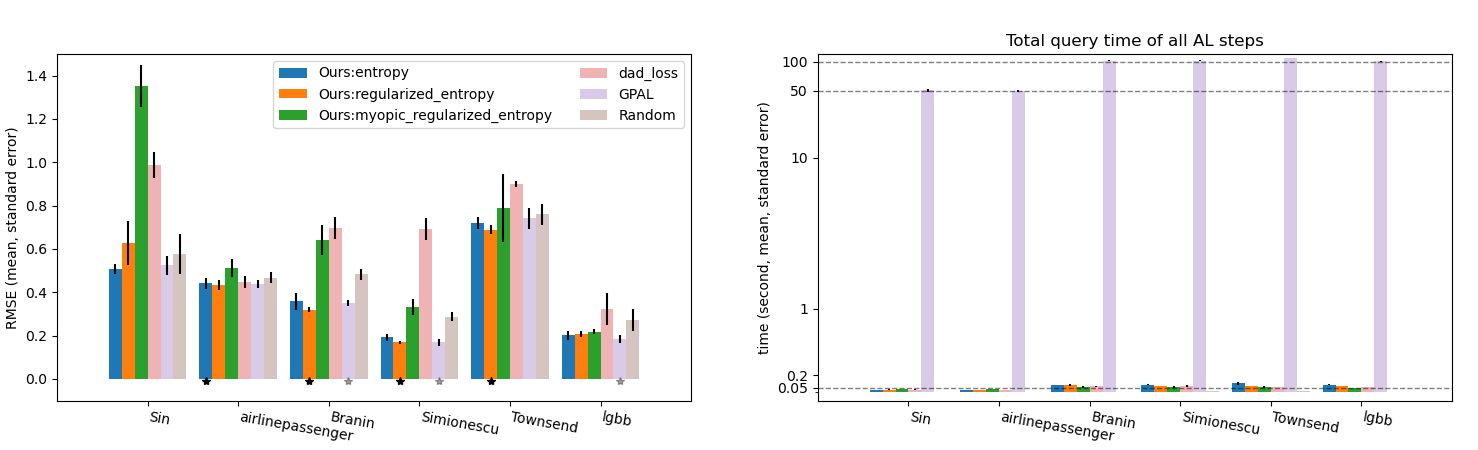}}
		\caption{
			A policy is trained with~\cref{alg-nonmyopic_al_training} together with the entropy or regularized entropy objective.
                See~\cref{table-training_time} for the training time.
			A myopic policy is trained with~\cref{alg-myopic_al_training}, where detail is given in~\cref{sectionS3-losses_details}.
			$N_{init}=1$.
			For 1D problems (Sin \& airline passenger), $T=10$, which means a total of $11$ observations.
			For 2D problems, $T=20$, which means a total of $21$ observations.
                For each benchmark problem, a star is marked if RMSE of the method is significantly smaller than Random (Wilcoxon signed-rank test, p-value threshold $0.05$).
			The time only takes querying time into account.
			For GP AL method, GP predictive distributions are necessary for the acquisition function and thus the GP training time is part of the querying time~\crefp{figure1,alg-classical_al}.
			Output labeling time is excluded.
		}
		\label{figure2}
	\end{center}
	\vskip -0.2in
\end{figure}

\begin{table}[b]
		\begin{center}
                \caption{Policy training time}\label{table-training_time}
			\begin{tabular}{r|ccc}
				\toprule
				\textbf{loss functions}
				&\textbf{entropy~\crefp{eq-gp_entropy_loss2_with_initial_points}}
				&\textbf{regularized entropy~\crefp{eq-gp_mi_loss2_grid_approx_with_initial_points}}
                    &\textbf{DAD baseline}\\
				training steps
				&$20k$
				&$10k$
				&$20k$\\
				\hline
				NN 1D, $T=10$
				  & 100 to 120 mins
				  & around 70 mins
				  & around 70 mins\\
				NN 2D, $T=20$
				  & 230 to 240 mins
				  & around 160 mins
				  & around 130 mins\\
				\bottomrule
			\end{tabular}\\
            The batch size is described in~\cref{sectionS5-numerical_details}.
            Regularized entropy method splits the batch into two and computes sub batches in sequence.
            The training is on one NVIDIA v100 (32GB GPU memory).
    \end{center}
\end{table}

In this section, we test our methods on a couple of benchmark tasks.

\subsection{NN training}

We prepare the experiments by running~\cref{alg-nonmyopic_al_training}, which corresponds to the up-front preparation block in~\cref{figure1}.
The entire~\cref{alg-nonmyopic_al_training} is one NN training step.
We implement the training pipeline with PyTorch.
In the following experiments, we train one NN policy for 1D benchmark tasks and one NN for 2D tasks.
The training time and the hardware are described in~\cref{table-training_time}.
The state dict (PyTorch model parameters) takes around $200$ KB disk space for both NNs.
The training of each setting is repeated five times with different random seeds.
Among the five training jobs of each NN, we select the NN with the best training loss for the following experiments.
See~\cref{sectionS-ablation} for details.

\subsection{Benchmark tasks}

We deploy AL over the following benchmark problems.
Our NNs are trained with $\mathcal{X}=[0, 1]^D$.

\paragraph{Sin function (1D):}

This is a one dimension problem $x \in [0, 1]$, $f(x)=\sin(20x)$.
In the experiments, we sample Gaussian noise $\epsilon \sim \mathcal{N}\left(0, 0.1^2\right)$.

\paragraph{Branin function (2D):}
This function is defined over $(x_1, x_2)=\in [-5, 10] \times [0, 15]$, which requires a rescaling mapping $\forall \bm{x} \in \mathcal{X}, (x_1, x_2)=(-5+15[\bm{x}]_1, 15[\bm{x}]_2)$.
The function
\begin{align*}
f_{a, b, c, r, s, t}\left((x_1, x_2)\right) &= a(x_2 - bx_1^2 + cx_1 -r) + s(1-t) cos(x_1) + s,
\end{align*}
where $a, b, c, r, s, t=(1, \frac{5.1}{4 \pi^2}, \frac{5}{\pi}, 6, 10, \frac{1}{8 \pi})$ are constants.
We sample noise free data points and use the samples to normalize our output
$
f_{a, b, c, r, s, t}\left((x_1, x_2)\right)_{normalize}
=\frac{
	f_{a, b, c, r, s, t}\left((x_1, x_2)\right) - mean(f_{a, b, c, r, s, t})
}{
	std(f_{a, b, c, r, s, t})
}.
$
In the experiments, the noise is $\epsilon \sim \mathcal{N}\left(0, 0.1^2\right)$.

\paragraph{Unconstrained Simionescu function (2D):}
This is originally a constrained problem~\citep{Simionescu_function} defined over $(x_1, x_2) \in [-1.25, 1.25]^2$ (which again requires a rescaling mapping $\mathcal{X}\rightarrow[-1.25, 1.25]^2$).
We remove the constraint, resulting in $f(x_1, x_2)=0.1 x_1 x_2$.
As Branin function, we sample noise free data points and use the samples to normalize our output.
In the experiments, the noise is $\epsilon \sim \mathcal{N}\left(0, 0.1^2\right)$.

\paragraph{Unconstrained Townsend function (2D):}
This is originally a constrained problem~\citep{Townsend_function}~\footnote{https://www.chebfun.org/examples/opt/ConstrainedOptimization.html} defined over $(x_1, x_2) \in [-2.25, 2.25]\times[-2.5, 1.75]$ (rescaling mapping from $\mathcal{X}$ required).
We remove the constraint, resulting in 
$
f(x_1, x_2)= - \left[ \cos((x_1-0.1)x_2) \right]^2 -x_1 \sin(3x_1 + x_2).
$
As Branin function, we sample noise free data points and use the samples to normalize our output.
In the experiments, the noise is $\epsilon \sim \mathcal{N}\left(0, 0.1^2\right)$.

\paragraph{Airline passenger dataset (1D):}
This is a publically available time series dataset~\footnote{https://github.com/jbrownlee/Datasets/blob/master/airline-passengers.csv}.
Each data point has a date input (year and month) and a number of passengers as output.
We convert the input into real number as $year + (month-1)/12$, and then rescale the entire input space to $[0, 1]$ (the earliest date becomes $0$ while the latest becomes $1$).
The output data are again normalized to zero mean and unit variance.

\paragraph{Langley Glide-Back Booster (LGBB) dataset (2D):}
This is a two dimension dataset described in~\cite{LGBB}\footnote{https://bobby.gramacy.com/surrogates/lgbb.tar.gz, lgbb\_original.txt}.
The dataset has multiple outputs and we take the "lift" to run our experiments (after normalized to zero mean and unit variance).
The inputs are $x_1$ (mach) and $x_2$ (alpha).
 which are normalized by
\begin{align*}
	x_1 &= mach / 6,\\
	x_2 &= (alpha + 5) / 35.
\end{align*}
After doing this, the input space is $[0, 1]^2$.

\subsection{AL deployment}
We compare our methods with (i) standard GP AL~\crefp{alg-classical_al} with entropy acquisition~\crefp{sectionS5-numerical_details} (ii) random selection criterion and (iii) DAD, i.e. amortized Bayesian experimental design proposed by~\cite{foster_dad_2021}.
In this section, we report the modeling performance and AL deployment time.
Since the high-level goal is to model a regression task, we use the collected datasets to train models and evaluate the RMSE as the modeling performance.
Although DAD and our amortized AL methods are not restricted to GP modeling, we still evaluate the data on GP models, as GPs are powerful modeling tools for such amount of data and as this is a fair comparison to baseline (i).

We run experiments over the aforementioned benchmark problems.
Our NN policy returns points on continuous space $\mathcal{X} \subseteq \mathbb{R}^D$.
On benchmark functions, a query is taken as it is (line 2 of~\cref{alg-policy_al}), while on the testing datasets (airline passenger and LGBB), we take the nearest point with $L2$-norm from the pool.
Notice that the single pre-trained 1D NN policy is used for all the 1D tasks and the 2D NN policy for all the 2D tasks.

For each method, we repeat the AL experiments~\crefp{alg-classical_al,alg-policy_al} for five times and report the mean and standard error.
Each experiment is executed with individual seed.
Note here that initial datasets (and noises of function problems) are randomly sampled, where the seed plays a role.

The results are shown in~\cref{figure2}.
The RMSEs are evaluated after the AL deployments.
For example, with 1 dim problems (sin \& airline passenger dataset), we start with $1$ initial points and query for $10$ iterations, resulting in $11$ data points in the end.
Then the RMSEs are evaluated with GPs trained with these $11$ data points.
The query time is the data selection time of all iterations.
We can see that, on all the presented benchmark problems except for the Sin function, data selected by our nonmyopic amortized AL approaches achieve as good modeling performances as conventional GP AL, while the querying time is significantly faster.
Some of the RMSE out-performance of our nonmyopic approaches (and the GP AL baseline) over Random is statistically significant (Wilcoxon signed-rank test, p-value smaller than $0.05$).
With myopic training scheme, the policy can perform well in some tasks such as the LGBB but badly in others.
In our~\cref{sectionS-ablation}, we present few more trained policies good at different tasks.
The DAD baseline sometimes performs well on $1D$ problems but not on any $2D$ problems.

In general, we consider this result as a huge success.
The tens-of-milliseconds-level decision-making time per query allows amortized AL method to be applied to systems where output responses are given in a few dozen Hz.
In such systems, it is obviously expensive to wait for GP modeling and entropy optimization.

\begin{acknowledgments}
This work was supported by Bosch Center for Artificial Intelligence, which provided financial support, computers and GPU clusters.
The Bosch Group is carbon neutral. Administration, manufacturing and research activities no longer leave a carbon footprint. This also includes GPU clusters on which the experiments have been performed.
\end{acknowledgments}

\bibliography{ref}

\begin{thebibliography}{36}
\expandafter\ifx\csname natexlab\endcsname\relax\def\natexlab#1{#1}\fi
\providecommand{\url}[1]{\texttt{#1}}
\providecommand{\href}[2]{#2}
\providecommand{\path}[1]{#1}
\providecommand{\DOIprefix}{doi:}
\providecommand{\ArXivprefix}{arXiv:}
\providecommand{\URLprefix}{URL: }
\providecommand{\Pubmedprefix}{pmid:}
\providecommand{\doi}[1]{\href{http://dx.doi.org/#1}{\path{#1}}}
\providecommand{\Pubmed}[1]{\href{pmid:#1}{\path{#1}}}
\providecommand{\bibinfo}[2]{#2}
\ifx\xfnm\relax \def\xfnm[#1]{\unskip,\space#1}\fi
\bibitem[{Settles(2010)}]{settles2010_al}
\bibinfo{author}{B.~Settles},
\newblock \bibinfo{title}{Active learning literature survey},
\newblock \bibinfo{journal}{University of Wisconsin-Madison}  (\bibinfo{year}{2010}).
\bibitem[{Kumar and Gupta(2020)}]{KumarGupta2020}
\bibinfo{author}{P.~Kumar}, \bibinfo{author}{A.~Gupta},
\newblock \bibinfo{title}{Active learning query strategies for classification, regression, and clustering: A survey},
\newblock \bibinfo{journal}{Journal of Computer Science and Technology}  (\bibinfo{year}{2020}).
\bibitem[{Tharwat and Schenck(2023)}]{tharwat2023_al}
\bibinfo{author}{A.~Tharwat}, \bibinfo{author}{W.~Schenck},
\newblock \bibinfo{title}{A survey on active learning: State-of-the-art, practical challenges and research directions},
\newblock \bibinfo{journal}{Mathematics}  (\bibinfo{year}{2023}).
\bibitem[{Gal et~al.(2017)Gal, Islam, and Ghahramani}]{Gal2017_icml_bal}
\bibinfo{author}{Y.~Gal}, \bibinfo{author}{R.~Islam}, \bibinfo{author}{Z.~Ghahramani},
\newblock \bibinfo{title}{Deep {B}ayesian active learning with image data},
\newblock \bibinfo{journal}{International Conference on Machine Learning}  (\bibinfo{year}{2017}).
\bibitem[{Kirsch et~al.(2019)Kirsch, van Amersfoort, and Gal}]{kirsch2019batchbald}
\bibinfo{author}{A.~Kirsch}, \bibinfo{author}{J.~van Amersfoort}, \bibinfo{author}{Y.~Gal},
\newblock \bibinfo{title}{Batchbald: Efficient and diverse batch acquisition for deep bayesian active learning},
\newblock \bibinfo{journal}{Advances in Neural Information Processing Systems}  (\bibinfo{year}{2019}).
\bibitem[{Lederer et~al.(2021)Lederer, Conejo, Maier, Xiao, Umlauft, and Hirche}]{Lederer2021_icml_gprealtimecontrol}
\bibinfo{author}{A.~Lederer}, \bibinfo{author}{A.~J.~O. Conejo}, \bibinfo{author}{K.~A. Maier}, \bibinfo{author}{W.~Xiao}, \bibinfo{author}{J.~Umlauft}, \bibinfo{author}{S.~Hirche},
\newblock \bibinfo{title}{Gaussian process-based real-time learning for safety critical applications},
\newblock \bibinfo{year}{2021}.
\bibitem[{Baram et~al.(2004)Baram, El-Yaniv, and Luz}]{Baram2004_jmlr_onlineAL}
\bibinfo{author}{Y.~Baram}, \bibinfo{author}{R.~El-Yaniv}, \bibinfo{author}{K.~Luz},
\newblock \bibinfo{title}{Online choice of active learning algorithms},
\newblock \bibinfo{journal}{Journal of Machine Learning Research}  (\bibinfo{year}{2004}).
\bibitem[{Konyushkova et~al.(2017)Konyushkova, Sznitman, and Fua}]{Konyushkova2017_learnAL}
\bibinfo{author}{K.~Konyushkova}, \bibinfo{author}{R.~Sznitman}, \bibinfo{author}{P.~Fua},
\newblock \bibinfo{title}{Learning active learning from data},
\newblock \bibinfo{journal}{Advances in Neural Information Processing Systems}  (\bibinfo{year}{2017}).
\bibitem[{Swersky et~al.(2020)Swersky, Rubanova, Dohan, and Murphy}]{Swersky2020amortized_bo}
\bibinfo{author}{K.~Swersky}, \bibinfo{author}{Y.~Rubanova}, \bibinfo{author}{D.~Dohan}, \bibinfo{author}{K.~Murphy},
\newblock \bibinfo{title}{Amortized bayesian optimization over discrete spaces},
\newblock \bibinfo{journal}{Conference on Uncertainty in Artificial Intelligence}  (\bibinfo{year}{2020}).
\bibitem[{Rasmussen and Williams(2006)}]{GPbook}
\bibinfo{author}{C.~Rasmussen}, \bibinfo{author}{C.~Williams},
\newblock \bibinfo{title}{Gaussian processes for machine learning},
\newblock \bibinfo{journal}{MIT Press}  (\bibinfo{year}{2006}).
\bibitem[{Guestrin et~al.(2005)Guestrin, Krause, and Singh}]{guestrin_mi_al_05}
\bibinfo{author}{C.~Guestrin}, \bibinfo{author}{A.~Krause}, \bibinfo{author}{A.~P. Singh},
\newblock \bibinfo{title}{Near-optimal sensor placements in gaussian processes},
\newblock \bibinfo{journal}{International Conference on Machine Learning}  (\bibinfo{year}{2005}).
\bibitem[{Krause and Guestrin(2007)}]{krause_nonmyopic_al_07}
\bibinfo{author}{A.~Krause}, \bibinfo{author}{C.~Guestrin},
\newblock \bibinfo{title}{Nonmyopic active learning of gaussian processes: An exploration-exploitation approach},
\newblock \bibinfo{journal}{International Conference on Machine Learning}  (\bibinfo{year}{2007}).
\bibitem[{Krause et~al.(2008)Krause, Singh, and Guestrin}]{krause08a}
\bibinfo{author}{A.~Krause}, \bibinfo{author}{A.~Singh}, \bibinfo{author}{C.~Guestrin},
\newblock \bibinfo{title}{Near-optimal sensor placements in gaussian processes: Theory, efficient algorithms and empirical studies},
\newblock \bibinfo{journal}{Journal of Machine Learning Research}  (\bibinfo{year}{2008}).
\bibitem[{Houlsby et~al.(2011)Houlsby, Huszar, Ghahramani, and Lengyel}]{houlsby_2011_BALD}
\bibinfo{author}{N.~Houlsby}, \bibinfo{author}{F.~Huszar}, \bibinfo{author}{Z.~Ghahramani}, \bibinfo{author}{M.~Lengyel},
\newblock \bibinfo{title}{Bayesian active learning for classification and preference learning},
\newblock \bibinfo{journal}{Computing Research Repository}  (\bibinfo{year}{2011}).
\bibitem[{Gershman and Goodman(2014)}]{Gershman2014amortized_inference}
\bibinfo{author}{S.~J. Gershman}, \bibinfo{author}{N.~D. Goodman},
\newblock \bibinfo{title}{Amortized inference in probabilistic reasoning},
\newblock \bibinfo{journal}{Annual Meeting of the Cognitive Science Society}  (\bibinfo{year}{2014}).
\bibitem[{Zimmer et~al.(2018)Zimmer, Meister, and Nguyen-Tuong}]{ZimmerNEURIPS2018_b197ffde}
\bibinfo{author}{C.~Zimmer}, \bibinfo{author}{M.~Meister}, \bibinfo{author}{D.~Nguyen-Tuong},
\newblock \bibinfo{title}{Safe active learning for time-series modeling with gaussian processes},
\newblock \bibinfo{journal}{Advances in Neural Information Processing Systems}  (\bibinfo{year}{2018}).
\bibitem[{Garnett et~al.(2014)Garnett, Osborne, and Hennig}]{GarnettOH2013}
\bibinfo{author}{R.~Garnett}, \bibinfo{author}{M.~Osborne}, \bibinfo{author}{P.~Hennig},
\newblock \bibinfo{title}{Active learning of linear embeddings for gaussian processes},
\newblock \bibinfo{journal}{Conference on Uncertainty in Artificial Intelligence}  (\bibinfo{year}{2014}).
\bibitem[{Schreiter et~al.(2015)Schreiter, Nguyen-Tuong, Eberts, Bischoff, Markert, and Toussaint}]{Schreiter2015}
\bibinfo{author}{J.~Schreiter}, \bibinfo{author}{D.~Nguyen-Tuong}, \bibinfo{author}{M.~Eberts}, \bibinfo{author}{B.~Bischoff}, \bibinfo{author}{H.~Markert}, \bibinfo{author}{M.~Toussaint},
\newblock \bibinfo{title}{Safe exploration for active learning with gaussian processes},
\newblock \bibinfo{journal}{Machine Learning and Knowledge Discovery in Databases}  (\bibinfo{year}{2015}).
\bibitem[{Yue et~al.(2021)Yue, Wen, Hunt, and Shi}]{Yue2021_al_gp_shape_control}
\bibinfo{author}{X.~Yue}, \bibinfo{author}{Y.~Wen}, \bibinfo{author}{J.~H. Hunt}, \bibinfo{author}{J.~Shi},
\newblock \bibinfo{title}{Active learning for gaussian process considering uncertainties with application to shape control of composite fuselage},
\newblock \bibinfo{journal}{IEEE Transactions on Automation Science and Engineering}  (\bibinfo{year}{2021}).
\bibitem[{Li et~al.(2022)Li, Rakitsch, and Zimmer}]{cyli2022}
\bibinfo{author}{C.-Y. Li}, \bibinfo{author}{B.~Rakitsch}, \bibinfo{author}{C.~Zimmer},
\newblock \bibinfo{title}{Safe active learning for multi-output gaussian processes},
\newblock \bibinfo{journal}{International Conference on Artificial Intelligence and Statistics}  (\bibinfo{year}{2022}).
\bibitem[{Bitzer et~al.(2023)Bitzer, Meister, and Zimmer}]{bitzer23a_nonstationary_al}
\bibinfo{author}{M.~Bitzer}, \bibinfo{author}{M.~Meister}, \bibinfo{author}{C.~Zimmer},
\newblock \bibinfo{title}{Hierarchical-hyperplane kernels for actively learning gaussian process models of nonstationary systems},
\newblock \bibinfo{year}{2023}.
\bibitem[{Seo et~al.(2000)Seo, Wallat, Graepel, and Obermayer}]{Seo2000GaussianPR}
\bibinfo{author}{S.~Seo}, \bibinfo{author}{M.~Wallat}, \bibinfo{author}{T.~Graepel}, \bibinfo{author}{K.~Obermayer},
\newblock \bibinfo{title}{Gaussian process regression: active data selection and test point rejection},
\newblock \bibinfo{journal}{Proceedings of the IEEE-INNS-ENNS International Joint Conference on Neural Networks. IJCNN 2000. Neural Computing: New Challenges and Perspectives for the New Millennium} \bibinfo{volume}{3} (\bibinfo{year}{2000}) \bibinfo{pages}{241--246 vol.3}. \URLprefix \url{https://api.semanticscholar.org/CorpusID:18551791}.
\bibitem[{Brochu et~al.(2010)Brochu, Cora, and de~Freitas}]{brochu2010tutorial}
\bibinfo{author}{E.~Brochu}, \bibinfo{author}{V.~M. Cora}, \bibinfo{author}{N.~de~Freitas},
\newblock \bibinfo{title}{A tutorial on bayesian optimization of expensive cost functions, with application to active user modeling and hierarchical reinforcement learning},
\newblock \bibinfo{journal}{arXiv}  (\bibinfo{year}{2010}).
\bibitem[{Rothfuss et~al.(2021)Rothfuss, Fortuin, Josifoski, and Krause}]{rothfuss21pacoh}
\bibinfo{author}{J.~Rothfuss}, \bibinfo{author}{V.~Fortuin}, \bibinfo{author}{M.~Josifoski}, \bibinfo{author}{A.~Krause},
\newblock \bibinfo{title}{Pacoh: Bayes-optimal meta-learning with pac-guarantees},
\newblock \bibinfo{journal}{International Conference on Machine Learning}  (\bibinfo{year}{2021}).
\bibitem[{Bitzer et~al.(2023)Bitzer, Meister, and Zimmer}]{Bitzer2023amortized_gp}
\bibinfo{author}{M.~Bitzer}, \bibinfo{author}{M.~Meister}, \bibinfo{author}{C.~Zimmer},
\newblock \bibinfo{title}{Amortized inference for gaussian process hyperparameters of structured kernels},
\newblock \bibinfo{journal}{Conference on Uncertainty in Artificial Intelligence}  (\bibinfo{year}{2023}).
\bibitem[{Andrychowicz et~al.(2016)Andrychowicz, Denil, G\'{o}mez, Hoffman, Pfau, Schaul, Shillingford, and de~Freitas}]{Andrychowicz2016_learn_to_learn}
\bibinfo{author}{M.~Andrychowicz}, \bibinfo{author}{M.~Denil}, \bibinfo{author}{S.~G\'{o}mez}, \bibinfo{author}{M.~W. Hoffman}, \bibinfo{author}{D.~Pfau}, \bibinfo{author}{T.~Schaul}, \bibinfo{author}{B.~Shillingford}, \bibinfo{author}{N.~de~Freitas},
\newblock \bibinfo{title}{Learning to learn by gradient descent by gradient descent},
\newblock \bibinfo{journal}{Advances in Neural Information Processing Systems}  (\bibinfo{year}{2016}).
\bibitem[{Chen et~al.(2017)Chen, Hoffman, Colmenarejo, Denil, Lillicrap, Botvinick, and de~Freitas}]{Chen2017_learn_to_learn}
\bibinfo{author}{Y.~Chen}, \bibinfo{author}{M.~W. Hoffman}, \bibinfo{author}{S.~G. Colmenarejo}, \bibinfo{author}{M.~Denil}, \bibinfo{author}{T.~P. Lillicrap}, \bibinfo{author}{M.~Botvinick}, \bibinfo{author}{N.~de~Freitas},
\newblock \bibinfo{title}{Learning to learn without gradient descent by gradient descent},
\newblock \bibinfo{journal}{International Conference on Machine Learning}  (\bibinfo{year}{2017}).
\bibitem[{Foster et~al.(2021)Foster, Ivanova, Malik, and Rainforth}]{foster_dad_2021}
\bibinfo{author}{A.~Foster}, \bibinfo{author}{D.~R. Ivanova}, \bibinfo{author}{I.~Malik}, \bibinfo{author}{T.~Rainforth},
\newblock \bibinfo{title}{Deep {Adaptive} {Design}: {Amortizing} {Sequential} {Bayesian} {Experimental} {Design}},
\newblock \bibinfo{journal}{International Conference on Machine Learning}  (\bibinfo{year}{2021}).
\bibitem[{Ivanova et~al.(2021)Ivanova, Foster, Kleinegesse, Gutmann, and Rainforth}]{ivanova_idad_2021}
\bibinfo{author}{D.~R. Ivanova}, \bibinfo{author}{A.~Foster}, \bibinfo{author}{S.~Kleinegesse}, \bibinfo{author}{M.~U. Gutmann}, \bibinfo{author}{T.~Rainforth},
\newblock \bibinfo{title}{Implicit {Deep} {Adaptive} {Design}: {Policy}-{Based} {Experimental} {Design} without {Likelihoods}}  (\bibinfo{year}{2021}).
\bibitem[{Rahimi and Recht(2007)}]{Rahimi2007_rff}
\bibinfo{author}{A.~Rahimi}, \bibinfo{author}{B.~Recht},
\newblock \bibinfo{title}{Random features for large-scale kernel machines},
\newblock \bibinfo{journal}{Advances in Neural Information Processing Systems}  (\bibinfo{year}{2007}).
\bibitem[{Wilson et~al.(2020)Wilson, Borovitskiy, Terenin, Mostowsky, and Deisenroth}]{Wilson2020_icml_rff_gp_post}
\bibinfo{author}{J.~T. Wilson}, \bibinfo{author}{V.~Borovitskiy}, \bibinfo{author}{A.~Terenin}, \bibinfo{author}{P.~Mostowsky}, \bibinfo{author}{M.~P. Deisenroth},
\newblock \bibinfo{title}{Efficiently sampling functions from gaussian process posteriors},
\newblock \bibinfo{journal}{International Conference on Machine Learning}  (\bibinfo{year}{2020}).
\bibitem[{Vaswani et~al.(2017)Vaswani, Shazeer, Parmar, Uszkoreit, Jones, Gomez, Kaiser, and Polosukhin}]{Vaswani2017_attention}
\bibinfo{author}{A.~Vaswani}, \bibinfo{author}{N.~Shazeer}, \bibinfo{author}{N.~Parmar}, \bibinfo{author}{J.~Uszkoreit}, \bibinfo{author}{L.~Jones}, \bibinfo{author}{A.~N. Gomez}, \bibinfo{author}{L.~u. Kaiser}, \bibinfo{author}{I.~Polosukhin},
\newblock \bibinfo{title}{Attention is all you need},
\newblock \bibinfo{journal}{Advances in Neural Information Processing Systems}  (\bibinfo{year}{2017}).
\bibitem[{Simionescu(2014)}]{Simionescu_function}
\bibinfo{author}{P.~Simionescu},
\newblock \bibinfo{title}{Computer-aided graphing and simulation tools for autocad users},
\newblock \bibinfo{journal}{Computer-Aided Graphing and Simulation Tools for AutoCAD Users}  (\bibinfo{year}{2014}).
\bibitem[{Townsend(2017)}]{Townsend_function}
\bibinfo{author}{A.~Townsend},
\newblock \bibinfo{title}{Constrained optimization in chebfun},
\newblock \bibinfo{journal}{chebfun.org}  (\bibinfo{year}{2017}).
\bibitem[{Rogers et~al.(2003)Rogers, Aftosmis, Pandya, Chaderjian, T., and Ahmad}]{LGBB}
\bibinfo{author}{S.~E. Rogers}, \bibinfo{author}{M.~J. Aftosmis}, \bibinfo{author}{S.~A. Pandya}, \bibinfo{author}{N.~M. Chaderjian}, \bibinfo{author}{E.~T. T.}, \bibinfo{author}{J.~U. Ahmad},
\newblock \bibinfo{title}{Automated cfd parameter studies on distributed parallel computers},
\newblock \bibinfo{journal}{AIAA Computational Fluid Dynamics Conference}  (\bibinfo{year}{2003}).
\bibitem[{Liu et~al.(2020)Liu, Jiang, He, Chen, Liu, Gao, and Han}]{Liu2020radam}
\bibinfo{author}{L.~Liu}, \bibinfo{author}{H.~Jiang}, \bibinfo{author}{P.~He}, \bibinfo{author}{W.~Chen}, \bibinfo{author}{X.~Liu}, \bibinfo{author}{J.~Gao}, \bibinfo{author}{J.~Han},
\newblock \bibinfo{title}{On the variance of the adaptive learning rate and beyond},
\newblock \bibinfo{journal}{International Conference on Learning Representations}  (\bibinfo{year}{2020}).

\end{thebibliography}

\newpage
\appendix

\renewcommand{\thepart}{Appendix}
\renewcommand{\partname}{}
\renewcommand{\ptctitle}{Overview}
\addcontentsline{toc}{section}{Appendix} 
\part{} 
\parttoc 

\section{Gaussian process and entropy}\label{sectionS2-gp_details}
We first write down the GP predictive distribution.
Given a set of $N_{observe}$ data points $\mathcal{D}=\{ \bm{X}, Y \} \subseteq \mathcal{X} \times \mathcal{Y}$, we wish to make inference at points $\bm{X}_{test} = \{ \bm{x}_{test,1}, ..., \bm{x}_{test,n} \}$.
We write $Y_{test}=(y(\bm{x}_{test,1}), ..., y(\bm{x}_{test,n}))$ for brevity.
The joint distribution of $Y$ and predictive $Y_{test}$ is Gaussian:
\begin{align}\label{eq-gp_prior_distribution}
\begin{split}
p(Y, Y_{test})
=\mathcal{N}\left(
\bm{0},
k_\theta( \bm{X} \cup \bm{X}_{test}, \bm{X} \cup \bm{X}_{test} )
+ \sigma^2 I_{N_{observe}+n}
\right)
\end{split}
\end{align}
where $k_\theta( \bm{X} \cup \bm{X}_{test}, \bm{X} \cup \bm{X}_{test} )$ is a gram matrix with $[k_\theta( \bm{X} \cup \bm{X}_{test}, \bm{X} \cup \bm{X}_{test} )]_{i,j}=k_\theta \left( [\bm{X} \cup \bm{X}_{test}]_i, [\bm{X} \cup \bm{X}_{test}]_j \right)$.

This leads to the following predictive distribution (or GP posterior distribution)
\begin{align}
\begin{split}\label{eq-gp_posterior}
p( Y_{test} | \mathcal{D})
&=\mathcal{N}\left(
Y_{test} \vert \mu_{\mathcal{D}}(\bm{X}_{test}), cov_{\mathcal{D}}(\bm{X}_{test})
\right),\\
\mu_{\mathcal{D}}(\bm{X}_{test})&=k_{\theta}\left( \bm{X}_{test}, \bm{X} \right)
\left[k_{\theta}\left( \bm{X}, \bm{X} \right) + \sigma^2 I_{N_{observe}}\right]^{-1} Y,\\
cov_{\mathcal{D}}(\bm{X}_{test})&=
k_{\theta}\left( \bm{X}_{test}, \bm{X}_{test} \right) + \sigma^2 I_n \\
&-
k_{\theta}\left( \bm{X}_{test}, \bm{X} \right)
\left[k_{\theta}\left( \bm{X}, \bm{X} \right) + \sigma^2 I_{N_{observe}}\right]^{-1}
k_{\theta}\left( \bm{X}, \bm{X}_{test} \right).
\end{split}
\end{align}
Elements of the predictive mean vector $\mu_{D}(\bm{X}_{test})$ are the noise-free predictive function values.

Note that the log probability density function is
\begin{align}
\begin{split}
\label{eq-gp_posterior_log_prob}
\log p( Y_{test} | \mathcal{D})=&
- 1/2 \log\left(
(2 \pi)^{n}
\det(cov_{\mathcal{D}}(\bm{X}_{test}))
\right)\\
&-1/2 (Y_{test} - \mu_{D}(\bm{X}_{test}))^T
\left[cov_{\mathcal{D}}(\bm{X}_{test})\right]^{-1}
(Y_{test} - \mu_{D}(\bm{X}_{test})),
\end{split}
\end{align}
and, if we consider $Y_{test}$ as a vector of $n$ random variables, the entropy is
\begin{align}
\label{eq-gp_posterior_entropy}
H( Y_{test} | \mathcal{D})=&
n/2 \log(2 \pi e) + 1/2 \log\det(cov_{\mathcal{D}}(\bm{X}_{test})).
\end{align}

Inverting a $N_{observe} \times N_{observe}$ matrix $\left[k_{\theta}\left( \bm{X}, \bm{X} \right) + \sigma^2 I_{N_{observe}}\right]$ has complexity $\mathcal{O}(N_{observe}^3)$ in time.
Computing the determinant also has cubic time complexity.

\section{Additional losses and myopic training algorithm}\label{sectionS3-losses_details}
In our main paper, we introduce
\begin{align*}
\mathcal{H}(\phi)\propto
&\mathbb{E}_{
	p(\theta, \sigma^2)
}
\mathbb{E}_{
	p(f(\cdot), \epsilon_{t=1, ...,T})
}
\left[
-\log p(y_{\phi,1}, ..., y_{\phi,T} | Y_{init})
\right],\\
\mathcal{I}(\phi)\propto
&\mathbb{E}_{
	p(\theta, \sigma^2)
}
\mathbb{E}_{
	p(f(\cdot), \epsilon_{t=1, ...,T})
} \left[
-\log p(y_{\phi,1}, ..., y_{\phi,T} | Y_{init})
+\log p( y_{\phi,1}, ..., y_{\phi,T} | Y_{init}, Y_{grid} )
\right].
\end{align*}

We additionally look into two more similar loss objectives.
We treat policy selected points as random variables, compute the entropy directly, and then take expectation over different priors and functions:
\begin{align}
\label{eq-gp_entropy_loss1_with_initial_points}
\mathcal{H}_{version2}(\phi) & = \mathbb{E}_{
	p(\theta, \sigma^2)
}
\mathbb{E}_{
	p(f(\cdot), \epsilon_{t=1, ...,T})
}
\left[
H(y_{\phi,1}, ..., y_{\phi,T} | Y_{init})
\right]\\
\label{eq-gp_mi_loss1_grid_approx_with_initial_points}
\mathcal{I}_{version2}(\phi)
&\approx \mathbb{E}_{
	p(\theta, \sigma^2)
}
\mathbb{E}_{
	p(f(\cdot), \epsilon_{t=1, ...,T})
} \left[
H(y_{\phi,1}, ..., y_{\phi,T} | Y_{init})
- H( y_{\phi,1}, ..., y_{\phi,T} | Y_{init}, Y_{grid} )
\right].
\end{align}

Substituting~\cref{eq-gp_posterior_log_prob,eq-gp_posterior_entropy} into the losses, we see that the key difference is whether the observation values $y_{\phi,1},...,y_{\phi,T}$ are taken into account.
We suspect that having $y_{\phi,1},...,y_{\phi,T}$ in the loss (main losses) may help the policy adapt in AL deployment.

Our ablation study below compares the losses.

\paragraph{Myopic policy training:}
\begin{wrapfigure}{R}{0.5\textwidth}
	\vspace{-5pt}
	\begin{minipage}{\linewidth}
		\captionof{algorithm}{Myopic AL training}
		\label{alg-myopic_al_training}
		\begin{algorithmic}[1]
			\Require prior $\mathcal{GP}(0, k_\theta), p(\epsilon)=\mathcal{N}(0, \sigma^2)$, $T$
			\State sample $\theta, \sigma^2$
			\State sample $f \sim \mathcal{GP}(0, k_\theta)$
			\State sample $t =1, ..., T$
			\State sample $\mathcal{D}_{t-1} \subseteq \mathcal{X}\times\mathcal{Y}$
			\State $\bm{x}_{t} = \phi(\mathcal{D}_{t-1})$
			\State sample $\epsilon_{t} \sim p(\epsilon), y_{t}=f(\bm{x}_{t}) + \epsilon_t$
			\State $\mathcal{D}_{t} \gets \mathcal{D}_{t-1} \cup \{ \bm{x}_{t}, y_{t} \}$
			\If{entropy loss}
			\State compute loss per~\cref{eq-gp_entropy_loss2_with_initial_points}
			\ElsIf{regularized entropy loss}
			\State sample $\bm{X}_{grid} \subseteq \mathcal{X}$
			\State sample $Y_{grid}=f(\bm{X}_{grid}) + noise$
			\State compute loss per~\cref{eq-gp_mi_loss2_grid_approx_with_initial_points}
			\EndIf
                \State update $\phi$
		\end{algorithmic}
	\end{minipage}
	\vspace{-5pt}
\end{wrapfigure}
As described in~\cref{section3-method}, we proposed another policy training method which does not require recursive NN forwarding.
The idea is simple: as the policy is intended to make AL with $N_{init}$ initial points for $T$ iterations, we sample the size of initial dataset from $N_{init}, ..., N_{init} + T - 1$ during the training, and then we simulate one-step AL.
This allows the policy to experience all sizes of datasets that it will be tackling during an AL deployment.
The training procedure is shown in~\cref{alg-myopic_al_training}.
All the loss functions are still the same: we condition on initial dataset with altered sizes, consider one-step query (compute as if $T=1$), and propagate the gradient.

\section{Training complexity}\label{sectionS4-training_complexity}

Overall, the training complexities are listed below.
\begin{itemize}
	\item computing loss: $\mathcal{O}\left( N_{init}^3 + T^3 \right)$ in time and $\mathcal{O}\left( N_{init}^2 + T^2 \right)$ in space for the entropy objective~\crefp{eq-gp_entropy_loss2_with_initial_points}, where the $N_{init}$ terms are time and cost of computing the GP predictive distribution (see~\cref{sectionS2-gp_details}) while the $T$ term of computing the log probability likelihood;
	\item computing loss: $\mathcal{O}\left( (N_{init}+N_{grid})^3 + T^3 \right)$ in time and $\mathcal{O}\left( (N_{init}+N_{grid})^2 + T^2 \right)$ in space for our regularized entropy objective~\crefp{eq-gp_mi_loss2_grid_approx_with_initial_points};
	\item computing loss: $\mathcal{O}\left( N_{init}^3 + T^3 \right)$ in time and $\mathcal{O}\left( N_{init}^2 + T^2 \right)$ in space for the entropy version 2 objective (appendix~\cref{eq-gp_entropy_loss1_with_initial_points});
	\item computing loss: $\mathcal{O}\left( (N_{init}+N_{grid})^3 + T^3 \right)$ in time and $\mathcal{O}\left( (N_{init}+N_{grid})^2 + T^2 \right)$ in space for our regularized entropy version 2 objective (appendix~\cref{eq-gp_mi_loss1_grid_approx_with_initial_points});
	\item NN forwarding: $\mathcal{O}\left( \sum_{t=1}^{T}(N_{init}+ t - 1)^2 \right)=\mathcal{O}\left( (N_{init} + T - 1)^3 - (N_{init} - 1)^3 \right)$ with the nonmyopic AL training~\crefp{alg-nonmyopic_al_training}, as self attention has square complexity~\cite{Vaswani2017_attention};
	\item NN forwarding: $\mathcal{O}\left( (N_{init}+ T - 1)^2 \right)$ with our myopic AL training~\crefp{alg-myopic_al_training} (this algorithm does not make recursive NN forwarding but the performance is worse).
\end{itemize}
Note that we train only once to get a policy for various AL problems.

\section{Numerical details}\label{sectionS5-numerical_details}

\paragraph{Policy training:}
In our current implementation, the data dimension and input bound need to be predefined.
We fix $\mathcal{X}=[0, 1]^D$, and rescale all test problems to this region.
The last layer of the NN policy is $layer(x)=(tanh(x) + 1)/2$ which ensures that the policy proposes points in $[0, 1]^D$.
The remaining structure is described in~\cite{ivanova_idad_2021}.

In~\cref{alg-nonmyopic_al_training,alg-myopic_al_training}, the kernel we use is a RBF kernel, which has $D+1$ variables: the variance $v$ and $D$ dimension lengthscale vector $\bm{l}$, i.e. $\theta=(v, \bm{l})$.
We sample $v \sim Uniform(0.505, 1.0)$, $\sigma^2=1.01 - v$ (function and noise variances sum to $1.01$) and $\bm{l} \sim Uniform(0.05, 1.0)$.
The sampling hyperparameters should be tuned according to the applications, our setting only use general assumptions.
The variance parameters utilize the assumptions that (i) data are normalized to unit variance and (ii) signal-to-noise ratio is at least one.
The lengthscale is kept general, but one has to make sure that it is numerically stable (e.g. too small is bad because each lengthscale component is a divisor in the kernel).

The GP functions are approximated by Fourier features, which means each function sample is a linear combination of cos functions~\citep{Rahimi2007_rff,Wilson2020_icml_rff_gp_post}:
\begin{align*}
f(\bm{x}) = \sum_{l=1}^{L} \omega_{l} \sqrt{2/L} \cos\left( \bm{a}_l^T\bm{x} + b_l \right),
\end{align*}
$\omega_l, b_l$ and $\bm{a}_l$ are sampled from distributions described in~\cite{Rahimi2007_rff}.
Larger $L$ leads to better approximations.
We set $L=100$.
The analytical mean of windows $[0, 1]^D$ is computed such that all functions can be shifted to zero mean (a GP function has zero mean and unit variance over the entire real space, but not necessarily in a specific bounded window).
The analytical mean is the integral of $f(x)$ divided by volume of $[0, 1]^D$.
The integral is
\begin{align*}
\frac{1}{a_l}\sum_{l=1}^{L} \omega_{l} \sqrt{2/L} &\sin\left( a_l x + b_l \right) \vert_{x=0}^1 \text{, for } D=1,\\
\frac{-1}{[\bm{a}_l]_1[\bm{a}_l]_2} \sum_{l=1}^{L} \omega_{l} \sqrt{2/L} &\cos\left( [\bm{a}_l]_1 x_1 + [\bm{a}_l]_2 x_2 + b_l \right) \vert_{x_1=0}^1\vert_{x_2=0}^1 \text{, for } D=2,\\
&\text{ and so on.}
\end{align*}
It may happen that at least one component of $\bm{a}_l$ is zero or is close to zero, which causes a problem in the division.
In this case, we replace $\vert 1 / \prod_{d=1}^D [\bm{a}_l]_d \vert$ by $100000$.
The error is negligible, i.e. much smaller than noise level.

The batch sizes are:
for nonmyopic training~\crefp{alg-nonmyopic_al_training}, we sample $25$ kernels ($25$ sets of $\theta$), $10$ sets of noise realizations $\epsilon_{t=1,...,T}$, $25$ functions per prior, resulting in overall $6250$ AL experiments per loss computation (expectation over $6250$ sequences of $T$ queries);
for myopic training~\crefp{alg-myopic_al_training}, we sample $250$ kernels, chunk them to $20$ different batches, each has its own size of initial datasets, and the remaining settings are the same.
The grid samples of regularized entropy objective have $N_{grid}=100$.
Note that whenever we need to sample input points, e.g. $\bm{X}_{init}, \bm{X}_{grid}$, we sample uniformly from $\mathcal{X}=[0, 1]^D$.

\paragraph{Experiments:}
In our experiments, we always model with a RBF kernel.
Given a dataset, GP hyperparameters are optimized with Type II maximum likelihood.

In our GP AL baseline, the acquisition function is the predictive entropy $H(y(\bm{x}) \vert \mathcal{D}_{t-1})$ (see~\cref{alg-classical_al} and~\cref{eq-gp_posterior_entropy}).
For the airline passenger and LGBB datasets, the acquisition score can be computed on the entire pool of unseen data, and then the optimization can be solved by selecting the point with the largest score.
For function problems, at each $t$, we randomly sample $5000$ inputs points, optimize on these points, make query and go to the next iterations.

\section{Ablation study}\label{sectionS-ablation}

\begin{figure}[t]
	\vskip 0.2in
	\begin{center}
		\centerline{\includegraphics[width=\textwidth]{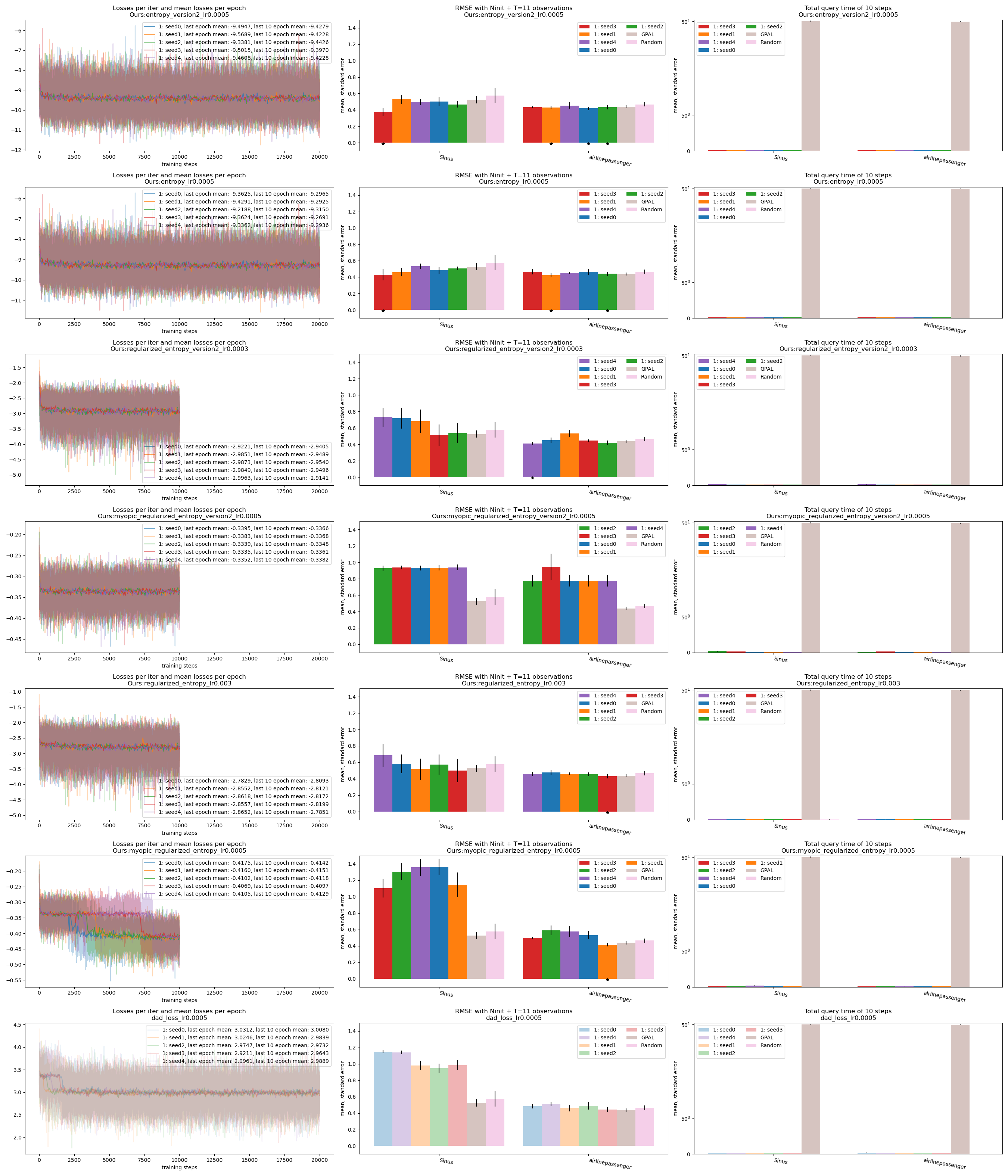}}
		\caption{
			A policy (row 1, 2, 3, 5) is trained with~\cref{alg-nonmyopic_al_training} together with the entropy or regularized entropy objective.
			A myopic policy (row 4) is trained with~\cref{alg-myopic_al_training}, where detail is given in~\cref{sectionS3-losses_details}.
			This plot show training of NN for 1D problems.
			The first column shows the \textbf{negative} loss values.
			In the second and third columns, we sort the trained policies by the mean training losses of the last 10 epochs (last 500 training steps).
			We set $N_{init}=1$ and $T=10$, which means a total of $11$ observations after AL is done.
                For each benchmark problem, a star is marked if RMSE of the method is significantly smaller than Random (Wilcoxon signed-rank test, p-value threshold $0.05$).
			The time only takes querying time into account.
			Output labeling time it excluded.
		}
		\label{figureS1D}
	\end{center}
	\vskip -0.2in
\end{figure}

\begin{figure}[t]
	\vskip 0.2in
	\begin{center}
		\centerline{\includegraphics[width=\textwidth]{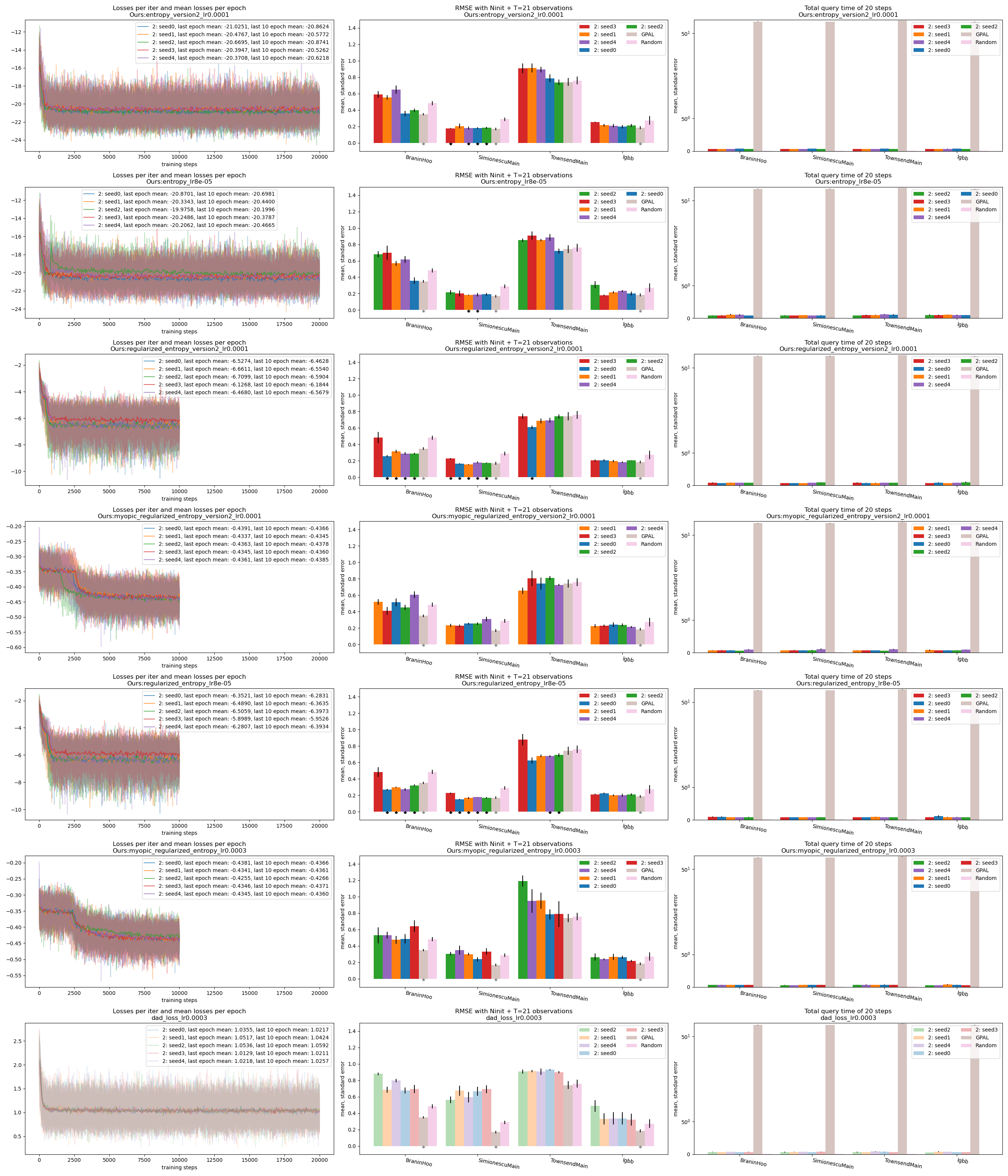}}
		\caption{
			A policy (row 1, 2, 3, 5) is trained with~\cref{alg-nonmyopic_al_training} together with the entropy or regularized entropy objective.
			A myopic policy (row 4) is trained with~\cref{alg-myopic_al_training}, where detail is given in~\cref{sectionS3-losses_details}.
			This plot show training of NN for 2D problems.
			The first column shows the \textbf{negative} loss values.
			In the second and third columns, we sort the trained policies by the mean training losses of the last 10 epochs (last 500 training steps).
			We set $N_{init}=1$ and $T=20$, which means a total of $21$ observations after AL is done.
                For each benchmark problem, a star is marked if RMSE of the method is significantly smaller than Random (Wilcoxon signed-rank test, p-value threshold $0.05$).
			The time only takes querying time into account.
			Output labeling time it excluded.
		}
		\label{figureS2D}
	\end{center}
	\vskip -0.2in
\end{figure}

\paragraph{Trained policy selection:}
For each training pipeline, we train with five different seeds.
The optimizer is RAdam~\citep{Liu2020radam}, and we try a few different initial learning rates (lrs).
We set a lr scheduler to discount the lr by $2\%$ every $50$ training steps.
With DAD objective, $\mathcal{H}$ and $\mathcal{H}_{version2}$, we train with $400*50=20000$ steps, and with $\mathcal{I}$ and $\mathcal{I}_{version2}$, we train with $200*50=10000$ steps.

The training results of $1$ dimension policy (our implementation pre-define dimensions) is shown in~\cref{figureS1D} and $2$ dimension in~\cref{figureS2D}.
With our main nonmyopic training, the training loss appears to be a good indicator of AL deployment performances.
Policies with the minimized negative objectives seem to perform the best in the test problems.
With myopic training, it seems like each trained policy may perform well in certain problems but badly in others.
In our main paper, for each training objective, we present the policy with best last-ten-epoch mean losses.

\end{document}